\documentclass[conference,a4paper]{IEEEtran}
\IEEEoverridecommandlockouts
\usepackage{cite}
\usepackage{amsmath,amssymb,amsfonts}
\usepackage{algorithmic}
\usepackage{graphicx}
\usepackage{textcomp}
\usepackage{xcolor}
\usepackage{comment}
\usepackage{multicol}
\usepackage{makecell}
\usepackage{float}
\usepackage{booktabs}
\usepackage{url}
\usepackage{eso-pic}
\def\BibTeX{{\rm B\kern-.05em{\sc i\kern-.025em b}\kern-.08em
    T\kern-.1667em\lower.7ex\hbox{E}\kern-.125emX}}
\begin{document}

\title{Self-Supervised Attention-Based Virtual Metrology for CMP MRR Prediction\\
\thanks{This research was supported by SUTD Research Surplus Grant No. RS-INSUR-00019-E0601-S00. (Corresponding authors: Nagarajan Raghavan and Senthilnath Jayavelu)}
}

\author{\IEEEauthorblockN{Jonathan Adam Rico}
\IEEEauthorblockA{\textit{Engineering Product Development,} 
\\
\textit{SUTD}\\
\textit{Institute of Advanced Intelligence and}\\
\textit{Computing, A*STAR}\\
Singapore \\
0000-0003-1023-6634}
\and
\IEEEauthorblockN{Nagarajan Raghavan}
\IEEEauthorblockA{\textit{Engineering Product Development,} 
\\
\textit{SUTD}\\
Singapore \\
0000-0001-6735-3108}
\and
\IEEEauthorblockN{Senthilnath Jayavelu}
\IEEEauthorblockA{\textit{Institute of Advanced Intelligence and} \\
\textit{Computing, A*STAR}\\
\textit{Dept. of Electrical and Computer} \\
\textit{Engineering, NUS}\\
Singapore \\
0000-0002-1737-7985}

}

\maketitle

\AddToShipoutPictureFG*{%
  \AtPageLowerLeft{%
    \raisebox{15mm}{%
      \parbox{0.95\paperwidth}{%
        \centering
        \footnotesize
        © 2026 IEEE.  Personal use of this material is permitted.  Permission from IEEE must be obtained for all other uses, in any current or future media, including reprinting/republishing this material for advertising or promotional purposes, creating new collective works, for resale or redistribution to servers or lists, or reuse of any copyrighted component of this work in other works.
      }%
    }%
  }%
}

\begin{abstract}
Chemical mechanical planarization (CMP) is a critical semiconductor manufacturing process where material removal rate (MRR) is a key metric for process monitoring and advanced process control (APC). Virtual metrology enables MRR estimation from process time series signals, supporting run-to-run (R2R) control and process optimization while reducing reliance on physical metrology. Existing approaches typically rely on simplified linear assumptions or require handcrafted feature engineering. In this paper, we propose Global Regression Pretraining (GREP), a self-supervised, attention-based framework for CMP MRR prediction. GREP learns temporal representations from unlabeled sensor signals through global proxy-target regression, followed by supervised fine-tuning for MRR estimation. The framework is evaluated against a feature-engineered supervised baseline under three conditions: limited labeled data, sensor noise, and temporal process drift. Experiments on the PHM 2016 CMP dataset demonstrate  competitive performance under limited labeled-data conditions, with improvements in selected cases and performance largely comparable to the baseline across different labeled-data fractions. Both models exhibit relatively small changes in prediction error with increasing noise, while GREP achieves a 17.0\% RMSE reduction in the final temporal interval.
\end{abstract}

\begin{IEEEkeywords}
Chemical mechanical planarization, Virtual metrology, Material removal rate, Self-supervised learning, Time series regression
\end{IEEEkeywords}

\section{Introduction}
Chemical mechanical planarization (CMP) is a critical process in semiconductor manufacturing, enabling surface planarization through combined chemical and mechanical interactions. Material removal rate (MRR) is a key process characteristic affecting wafer uniformity, process consistency, and manufacturing quality. Classical physics-based approaches, such as the Preston model \cite{preston1927}, provide foundational relationships between polishing conditions and material removal. However, accurate MRR estimation remains challenging because CMP involves complex interactions among process conditions, consumable states, and equipment behavior \cite{data_driven_mrr}.

Virtual metrology (VM) provides an alternative approach to estimating wafer quality characteristics using process measurements without requiring direct physical metrology for every wafer. Previous studies have demonstrated the potential of VM for semiconductor process monitoring and advanced process control (APC) \cite{cai2020}, \cite{virtual_metrology}. For CMP, physics-informed feature engineering and conventional machine learning have been investigated for MRR prediction \cite{data_driven_mrr}, \cite{li_baseline_2018}. Recently, Chen et al. \cite{chen_tabpfn_2026} used 20 selected features from 68 engineered descriptors with an externally pretrained TabPFN model, achieving an RMSE of 4.98 nm/min without hyperparameter tuning. However, such feature engineering requires domain knowledge and may discard temporal information from the original sensor signals.

Data-driven VM also faces several practical challenges. First, labeled metrology data are relatively costly to obtain, while equipment sensor signals are routinely collected during processing. Second, sensor noise and measurement disturbances may affect prediction accuracy. Third, changes in equipment and process conditions can introduce temporal distribution shifts, reducing the reliability of models trained on historical data \cite{virtual_metrology}, \cite{maitra_review_2024}. These considerations motivate models that can learn meaningful temporal representations with limited labeled data while maintaining predictive performance under changing operating conditions.

Self-supervised learning enables representation learning from unlabeled time series. Existing methods include contrastive learning \cite{ts_contrastive} and masked reconstruction \cite{ti_mae}. An alternative is to construct regression targets from predefined statistical, spectral, and temporal features, although this requires explicit feature selection and extraction. Random projections offer computationally inexpensive transformations of high-dimensional data \cite{achlioptas2003} and have recently been investigated as self-supervised learning targets. Sui et al. \cite{sui2024} employed random neural-network projectors for representation learning, while \cite{gui2024} used random-projection quantization for masked EEG time series prediction. Unlike these approaches, the proposed GREP directly performs regression on continuous global projections of complete and partial sensor sequences, without requiring predefined feature categories or target quantization.

In this work, we propose Global Regression Pretraining (GREP), a self-supervised, attention-based virtual metrology framework for CMP MRR prediction. GREP learns temporal representations by regressing global proxy targets constructed from unlabeled process signals, followed by supervised adaptation for MRR estimation. We investigate three research questions: (RQ1) whether GREP improves prediction accuracy with limited labeled data; (RQ2) how GREP performs under sensor noise; and (RQ3) how GREP performs under temporal process drift. These questions are evaluated using the PHM 2016 CMP dataset.

\section{GREP Framework}

The proposed GREP framework consists of self-supervised pretraining followed by supervised MRR regression, as illustrated in Fig. 1. Given a multivariate CMP sensor sequence \(\mathbf{X}_i \in \mathbb{R}^{T \times C}\), where \(T\) and \(C\) denote the sequence length and number of sensor channels, respectively, the input is divided into overlapping temporal patches and processed by a channel-independent Transformer encoder.

\begin{figure*}[htbp]
    \centering
    \includegraphics[width=0.95\linewidth]{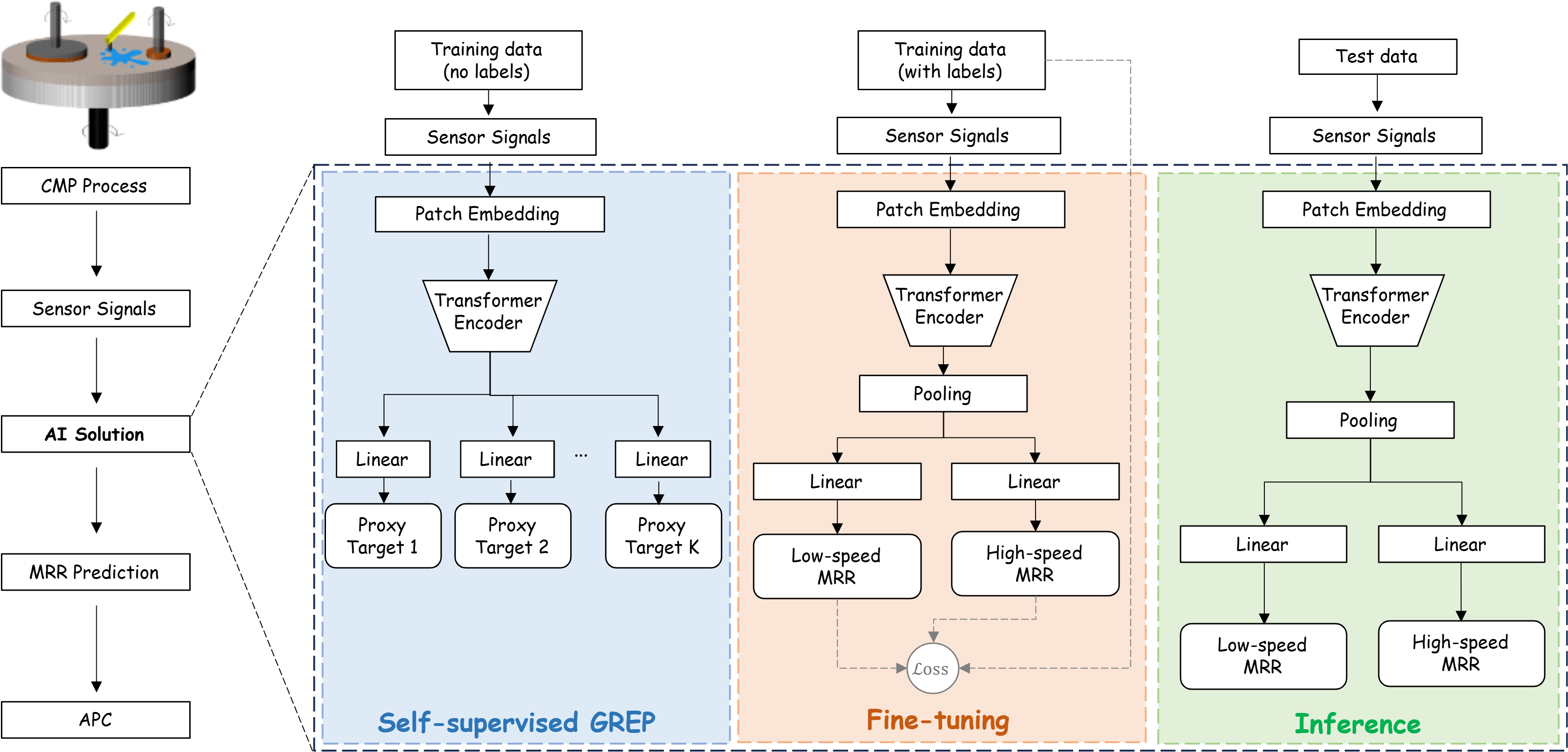}
    \caption{Overview of the proposed self-supervised GREP architecture for MRR virtual metrology.}
    \label{fig:grep_architecture}
\end{figure*}

\subsection{Self-Supervised Global Regression Pretraining}

During pretraining, each sensor channel is standardized and divided into three temporal views: the complete sequence, first half, and second half. Each view is projected onto 64 fixed Gaussian random vectors. The $k$-th proxy target from view (v) is defined by

\begin{equation}
    g_{i,c,k}^{(v)}
    =
    \sum_{t=1}^{T_v}
    \frac{x_{i,t,c}^{(v)}-\mu_{i,c}^{(v)}}
    {\sigma_{i,c}^{(v)}+\epsilon}
    W_{t,k},
    \quad W_{t,k}\sim\mathcal{N}(0,1)
    \label{eq:proxy}
\end{equation}
where $x$ is the input sensor sequence, $T_v$ denotes the view length, $\mu_{i,c}^{(v)}$ and $\sigma_{i,c}^{(v)}$ are its temporal mean and standard deviation, $\epsilon$ is a small constant, and $W_{t,k}\sim\mathcal{N}(0,1)$ is a fixed projection coefficient. Concatenating the three views produces $K=192$ proxy targets per channel.

The encoder and auxiliary regression heads are trained to predict these targets by minimizing the standardized mean squared error defined as

\begin{equation}
    \mathcal{L}_{\mathrm{GREP}} =
    \frac{1}{NCK}
    \sum_{i=1}^{N}\sum_{c=1}^{C}\sum_{k=1}^{K}
    \left(
    \frac{\hat{g}_{i,c,k}-g_{i,c,k}}
    {\sigma_k+\epsilon}
    \right)^2
    \label{eq:grep}
\end{equation}
where $N$ is the number of samples, $\hat{g}$ denotes the predicted pretraining target, and $\sigma_k$ is the standard deviation of the target batch-wise. This enables the model to learn global temporal representations without requiring measured MRR labels.

\subsection{Downstream MRR Regression}

Following pretraining, the auxiliary regression heads are replaced with MRR prediction heads. The pretrained encoder representations are aggregated using the final and mean-pooled patch features, followed by channel aggregation. Separate models are initialized with GREP-pretrained weights and fine-tuned for low-speed and high-speed CMP operation by minimizing the mean squared error between predicted and measured MRR. Both the encoder and downstream regression parameters are optimized using labeled training data.

\section{Experimental Methodology}

\subsection{CMP PHM 2016 Dataset and Preprocessing}

Experiments were conducted using the PHM 2016 CMP dataset \cite{cmp_dataset}, which contains 19 process variables, including pressure, flow rate, rotational speed, and usage counts. The prediction target is average MRR, expressed in nm/min. The data comprise two operating modes: low-speed polishing in chambers 4-6 and high-speed polishing in chambers 1-3. Table \ref{tab:cmp_modes} summarizes the data distribution for both operating modes. 

Sensor sequences were smoothed using a five-point moving average and linearly resampled to 128 time steps. Each sequence was divided into overlapping patches of length 16 and stride 8.

GREP was implemented in PyTorch using a channel-independent PatchTST encoder \cite{patchtst2023}. The Transformer comprised six encoder layers, eight attention heads, a hidden dimension of 128, and a feedforward dimension of 256. Gaussian Error Linear Unit (GELU) activation and a dropout rate of 0.1 were used. Self-supervised pretraining was performed for 150 epochs using AdamW, a learning rate of 0.001, and a batch size of 32. The pretraining stage used only the designated training partition.

For downstream regression, separate prediction models were fine-tuned for low-speed and high-speed operation. Learning rates of 0.001 and 0.0005 and batch sizes of 32 and 8 were used, respectively. Both models were fine-tuned for 50 epochs with mean squared error loss and a dropout rate of 0.4. The learning rate followed a five-epoch linear warmup and cosine annealing schedule. Experiments were conducted on an NVIDIA RTX 3070 GPU.

\begin{table}[tb]
    \centering
    \small
    \caption{Details on the number of train and test samples for each of the operating modes of a CMP run.}
    \begin{tabular}{c|cccc}
        \Xhline{2\arrayrulewidth}
        \noalign{\vskip 2pt}
        Mode & Chamber & Average MRR & Train & Test\\
        \noalign{\vskip 2pt}
        \hline
        \noalign{\vskip 2pt}
        Low-speed & 4,5,6 &55-110 nm/min & 1613 & 364\\
        High-speed & 1,2,3 &140-170 nm/min & 351 & 73\\
        \noalign{\vskip 2pt}
        \Xhline{2\arrayrulewidth}
    \end{tabular}
    \label{tab:cmp_modes}
\end{table}

\subsection{Research Questions and Evaluation}

The proposed method was compared with a supervised, feature-engineered baseline implemented following the approach of Li et al. \cite{li_baseline_2018}. Process-level features, including summary statistics, integrated sensor measurements, polishing duration, and pressure-related variables, were extracted from each CMP run. Separate tree-based regression models were trained for the two polishing-speed modes. Both methods were evaluated using the same experimental data partitions and labeled-data fractions. Root mean squared error (RMSE) was used as the evaluation metric, with lower values indicating better prediction accuracy.

\textbf{RQ1: Label Efficiency}. To assess performance with limited ground-truth measurements, GREP was pretrained on unlabeled training sequences, while downstream regression used 10\%-90\% of the labeled training data. Prediction accuracy was compared with the supervised baseline across the same labeled-data fractions.

\textbf{RQ2: Robustness to Sensor Noise}. Following the time-domain perturbation methods described in \cite{survey_wen2021}, Gaussian noise, random spikes, and linear trends were introduced into the training sensor signals. Gaussian noise was generated at SNR levels of 0, 5, 10, 20, and 30 dB. Spikes were applied independently with a probability of 0.01 and a standard deviation three times that of the Gaussian-perturbed signals. For each wafer-stage sequence, a linear trend with a randomly sampled terminal offset from $\mathcal{N}(0,0.05^2)$ was also added. These settings provide controlled synthetic perturbations rather than calibrated CMP sensor fault conditions. Models were evaluated on unmodified test data.

\textbf{RQ3: Robustness to Process Drift}. To assess robustness to temporal process changes, the training data were restricted to the earliest chronological interval, and prediction errors were evaluated over subsequent intervals. This setup examines the ability of models trained on earlier CMP observations to maintain accuracy as the temporal separation from the training period increases.

\section{Results and Discussion}

\subsection{RQ1: Label Efficiency}

Fig. \ref{fig:noise} compares GREP and the feature-engineered baseline across labeled-data fractions and training-signal noise levels. GREP achieves lower prediction errors in selected limited-label settings while maintaining comparable performance across most labeled-data fractions. Although the baseline performs notably better in one setting, the results demonstrate that GREP remains competitive under limited metrology data availability.


\subsection{RQ2: Robustness to Sensor Noise}

\begin{figure}[htbp]
    \centering
    \includegraphics[width=0.995\linewidth]{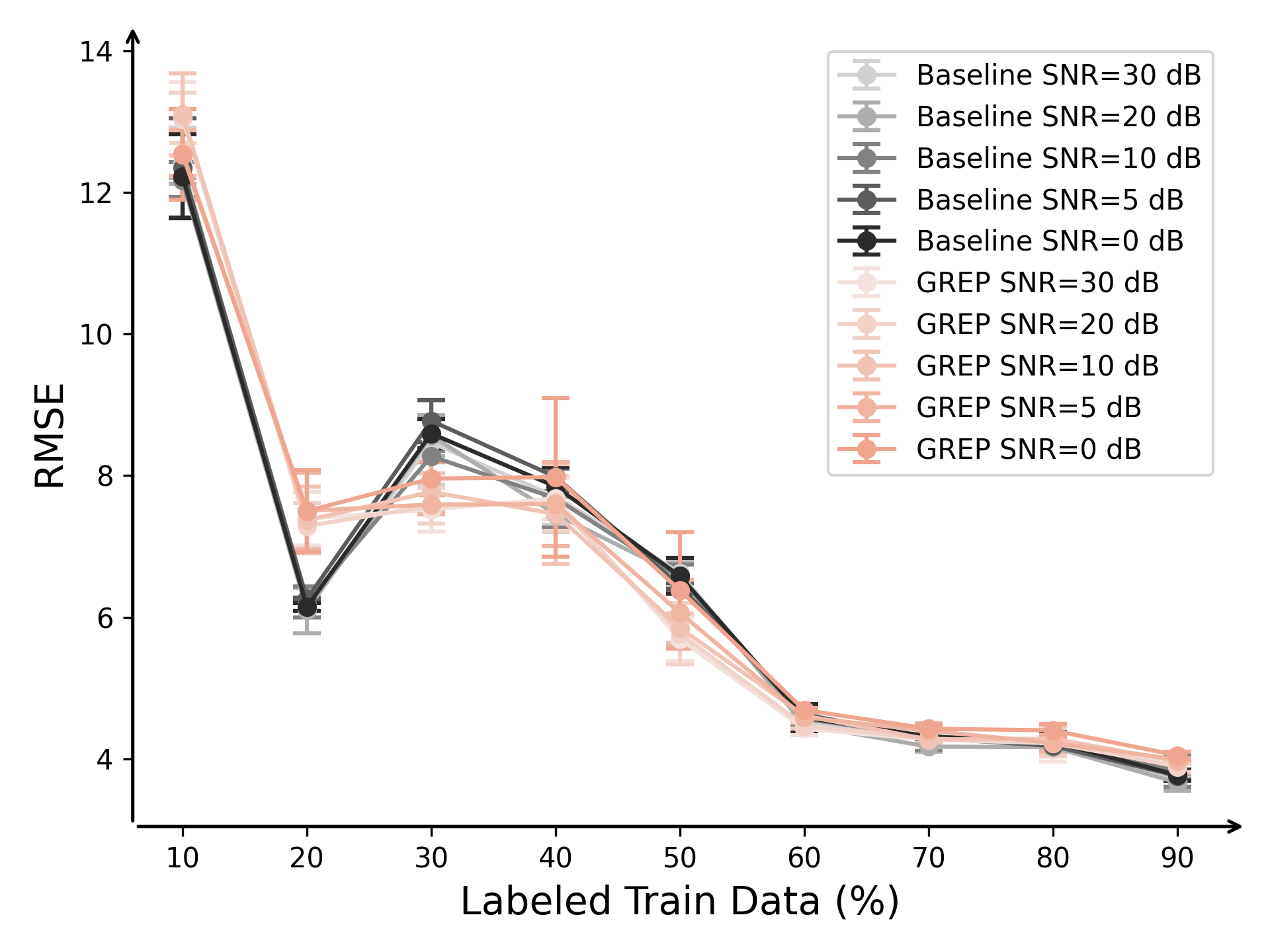}
    \caption{MRR Prediction RMSE under different training-signal noise levels and varying proportions of labeled training data.}
    \label{fig:noise}
\end{figure}

Noise sensitivity was assessed from the change in RMSE with increasing noise. Fig. \ref{fig:noise} shows that at 30\% labeled data, the fitted slopes were 0.114 and 0.049 RMSE per tenfold increase in noise power for GREP and the baseline, respectively. At 40\%, the corresponding slopes are 0.054 and 0.099. The fitted slopes remain relatively small, but neither method consistently exhibits lower noise sensitivity across labeled-data fractions. Thus, GREP maintains competitive prediction accuracy under training-signal corruption without demonstrating uniformly higher resistance to increasing noise intensity.

\subsection{RQ3: Robustness to Process Drift}

\begin{figure}[htbp]
    \centering
    \includegraphics[width=0.995\linewidth]{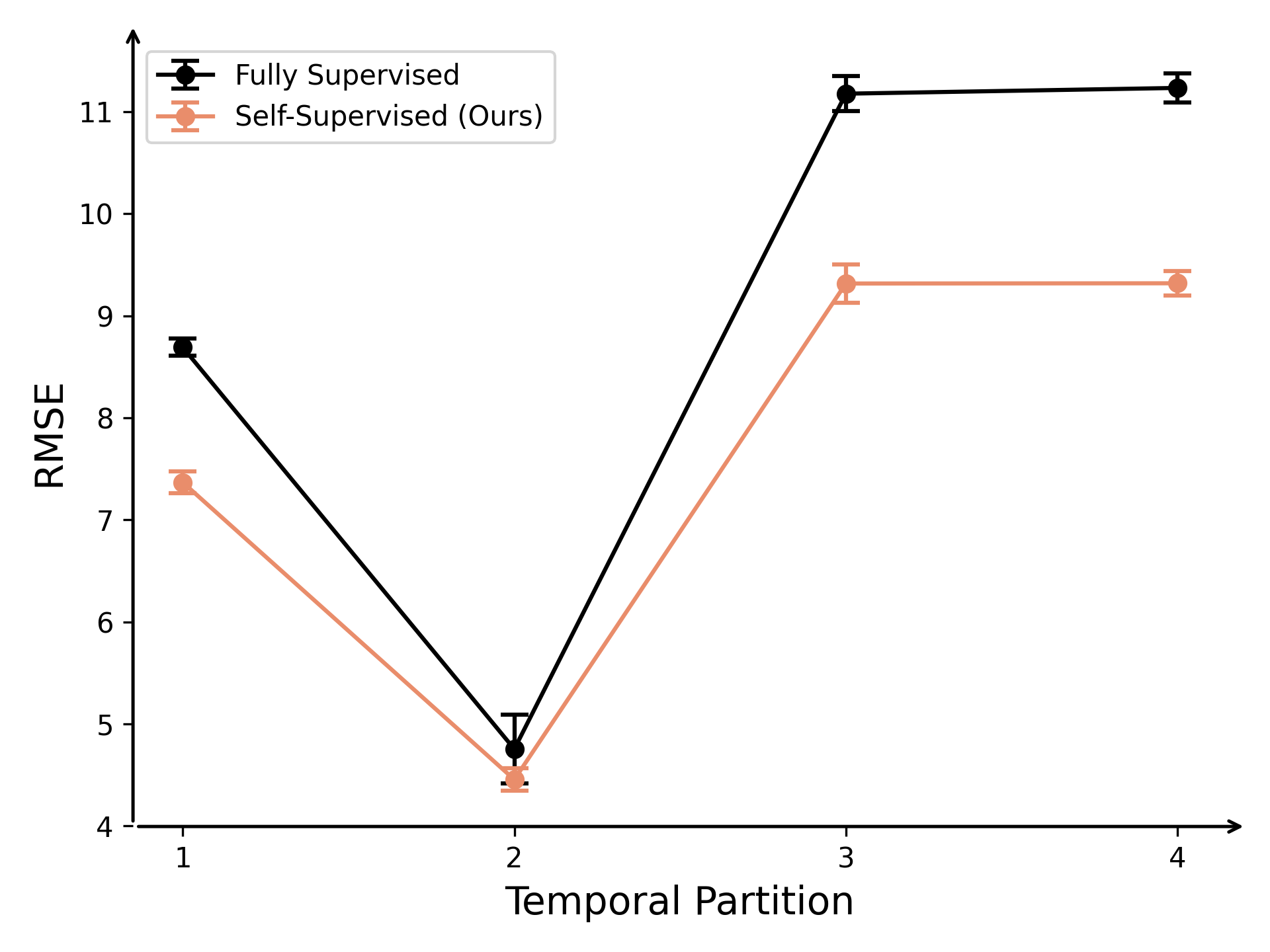}
    \caption{MRR prediction RMSE across four temporal partitions, with training restricted to the first partition.}
    \label{fig:process_drift}
\end{figure}

 Fig. \ref{fig:process_drift} presents prediction errors across four chronological intervals. GREP achieves lower RMSE than the baseline in all intervals, with reductions of 16.6\% and 17.0\% in the third and fourth intervals, respectively. In the final interval, GREP achieves an RMSE of 9.32, compared with 11.23 for the baseline.

The smaller prediction errors in later intervals suggest that GREP generalizes better to temporally separated CMP observations. Although the physical causes of process drift were not independently identified, the results indicate improved robustness to temporal process changes.

\section{Conclusion}

This study presented GREP, a self-supervised, attention-based virtual metrology framework for CMP MRR prediction. Compared with a feature-engineered supervised baseline, GREP achieved competitive accuracy under limited-label conditions and lower prediction errors across temporally separated CMP observations. Both models exhibited relatively small changes in prediction error under increasing training-signal noise, without a consistent advantage in noise sensitivity. These findings suggest that GREP can improve label-efficient MRR prediction and robustness to temporal process changes. Although APC integration is beyond the scope of this work, GREP could support run-to-run control and CMP process monitoring. Future work will investigate online adaptation to process drift, handcrafted versus proxy target ablation, and uncertainty-aware metrology sampling to evaluate the potential for reducing physical measurement requirements while maintaining prediction accuracy.

\section*{Acknowledgment}

The first author acknowledges the Singapore University of Technology and Design (SUTD) and the Ministry of Education, Singapore (MOE), for support through the Research Student Scholarship and SUTD Ph.D. Fellowship. This research was also supported by the Research Surplus Grant No. RGSUR08 of Prof. Nagarajan Raghavan. The authors thank the Agency for Science, Technology and Research (A*STAR) for providing computational resources.

\section*{Statement on use of Generative AI}

The authors acknowledge the use of ChatGPT (OpenAI) for grammar refinement and clarity improvement. All AI-assisted content was reviewed and edited by the authors to ensure accuracy and compliance with IEEE publication standards.

\bibliographystyle{IEEEtran}
\bibliography{references}

@article{preston1927,
    author = {F.W. Preston},
    title = {The theory and design of plate glass polishing machines},
    journal = {Journal of the Society of Glass Technology},
    year = {1927}
}

@inproceedings{li_baseline_2018,
    author = {X. Li and C. Wang and X. Mo and D. Zhao and C. Li},
    title = {Assessment of physics-based and data-driven models for material removal rate prediction in chemical mechanical polishing},
    booktitle = {2nd International Conference on Electrical Engineering and Automation},
    year = {2018},
    pages = {116-121}
}

@misc{cmp_dataset,
    author = {{Society of Prognostics and Health Management}},
    title = {{PHM Data Challenge}},
    year = {2016},
    note = {Dataset},
    howpublished = {[Online]. Available: \url{https://www.phmsociety.org/events/conference/phm/16/data-challenge}}
}

@article{chen_tabpfn_2026,
    author = {W. Chen and X. Han and Z. Wu and Z. Zhu and Y. Peng and J. Fu and S. Yuan and D. Yang},
    title = {Physics-Guided Feature Construction for Small-Sample Virtual Metrology in Chemical Mechanical Polishing},
    journal = {IEEE Transactions on Semiconductor Manufacturing},
    year = {2026}
}

@inproceedings{patchtst2023,
    author = {Y. Nie and N. Nguyen and P. Sinthong and J. Kalagnanam},
    title = {A time series is worth 64 words: Long-term forecasting with transformers},
    booktitle = {International Conference on Learning Representations},
    year = {2023}
}

@article{virtual_metrology,
    author = {A. Khan and J. Moyne and D. Tilbury},
    title = {Virtual metrology and feedback control for semiconductor manufacturing processes using recursive partial least squares},
    journal = {Journal of Process Control},
    year = {2008},
    volume = {18},
    issue = {10},
    pages = {961-974}
}

@inproceedings{data_driven_mrr,
    author = {Z. Li and D. Wu},
    title = {A Data-driven approach to material removal rate prediction in chemical mechanical polishing},
    booktitle = {Annual Conference of the PHM Society},
    year = {2018},
    volume = {10},
    issue = {1}
}

@inproceedings{ts_contrastive,
    author = {E. Eldele and M. Ragab and Z. Chen and M. Wu and C. Kwoh and X. Li and C. Guan},
    title = {Time-Series Representation Learning via Temporal and Contextual Contrasting},
    booktitle = {Proceedings of the Thirtieth International Joint Conference on Artificial Intelligence},
    year = {2021},
    pages = {2352-2359}
}

@unpublished{ti_mae,
    author = {Z. Li and Z. Rao and L. Pan and P. Wang and Z. Xu},
    title = {Ti-mae: Self-supervised masked time series autoencoders},
    note = {arXiv preprint arXiv:2301.08871}
}

@article{maitra_review_2024,
    author = {V. Maitra and Y. Su and J. Shi},
    title = {Virtual metrology in semiconductor manufacturing: Current status and future prospects},
    journal = {Expert Systems with Applications},
    year = {2024},
    page = {123559},
    volume = {249}
}

@article{cai2020,
    author = {H. Cai and J. Feng and Q. Yang and W. Li and X. Li and J. Lee},
    title = {A virtual metrology method with prediction uncertainty based on Gaussian process for chemical mechanical planarization},
    journal = {Computers in Industry},
    year = {2020},
    volume = {119},
    page = {103228}    
}

@article{survey_wen2021,
    author = {Q. Wen and L. Sun and F. Yang and X. Song and J. Gao and X. Wang and H. Xu},
    title = {Time Series Data Augmentation for Deep Learning: A Survey},
    journal = {International Joint Conference on Artificial Intelligence},
    year = {2021},
    pages = {4653–4660}
}

@inproceedings{sui2024,
    author = {Y. Sui and T. Wu and J. Cresswell and G. Wu and G. Stein and X.S.G. Huang and X. Zhang and M. Volkovs},
    title = {Self-supervised representation learning from random data projectors},
    booktitle = {International Conference on Learning Representations},
    year = {2024},
    pages = {6792-6814}
}

@article{achlioptas2003,
    author = {Achlioptas, D.},
    title = {Database-friendly random projections: Johnson-Lindenstrauss with binary coins},
    journal = {Journal of Computer and System Sciences},
    year = {2003},
    volume = {66},
    issue = {4},
    pages = {671-687}
}

@inproceedings{gui2024,
    author = {Gui, H. and Li, X. and Chen, X.},
    title = {Vector Quantization Pretraining for EEG Time Series with Random Projection and Phase Alignment},
    booktitle = {International Conference on Machine Learning},
    year = {2024},
    pages = {16731-16750}
}

\section*{Author Biography}

The first author is a PhD candidate at the SUTD with research attachment at the A*STAR, Singapore. His research focuses on time series foundation models for industrial applications. 

\end{document}